\documentclass{article}

\PassOptionsToPackage{numbers, compress}{natbib}

\usepackage[preprint]{neurips_2024}

\usepackage[utf8]{inputenc} 
\usepackage[T1]{fontenc}    
\usepackage{hyperref}       
\usepackage{url}            
\usepackage{booktabs, cellspace}       
\usepackage{amsfonts}       
\usepackage{nicefrac}       
\usepackage{microtype}      
\usepackage{xcolor}         

\usepackage{amsmath,amssymb}
\usepackage{hyperref}
\usepackage{graphicx}
\usepackage{multirow}
\usepackage{makecell}
\usepackage[cjk]{kotex}
\usepackage{caption}
\usepackage{wrapfig}
\usepackage{xspace}
\usepackage{color}
\usepackage{subfig}
\usepackage[capitalize]{cleveref}
\renewcommand\footnotemark{}

\hypersetup{
    colorlinks=true,
    citecolor=teal,
    urlcolor=teal
}

\makeatletter
\newcommand\figcaption{\def\@captype{figure}\caption}
\newcommand\tabcaption{\def\@captype{table}\caption}
\makeatother

\title{Depth-Guided Semi-Supervised Instance Segmentation}

\author{%
  \textbf{Xin Chen$^1$, Jie Hu$^2$, Xiewu Zheng$^1$, Jianghang Lin$^1$, Liujuan Cao$^{1*}$\thanks{*Corresponding author}, Rongrong Ji$^1$} \\
  $^1$Key Laboratory of Multimedia Trusted Perception and Efficient Computing, \\
  Ministry of Education of China, Xiamen University \\
  $^2$Contemporary Amperex Technology Co. Limited \\ 
}

\begin{document}

\maketitle

\begin{abstract}
Semi-Supervised Instance Segmentation (SSIS) aims to leverage an amount of unlabeled data during training. Previous frameworks primarily utilized the RGB information of unlabeled images to generate pseudo-labels. However, such a mechanism often introduces unstable noise, as a single instance can display multiple RGB values.
To overcome this limitation, we introduce a Depth-Guided (DG) SSIS framework.
This framework uses depth maps extracted from input images, which represent individual instances with closely associated distance values, offering precise contours for distinct instances. 
Unlike RGB data, depth maps provide a unique perspective, making their integration into the SSIS process complex.
To this end, we propose Depth Feature Fusion, which integrates features extracted from depth estimation. This integration allows the model to understand depth information better and ensure its effective utilization. 
Additionally, to manage the variability of depth images during training, we introduce the Depth Controller. This component enables adaptive adjustments of the depth map, enhancing convergence speed and dynamically balancing the loss weights between RGB and depth maps.
Extensive experiments conducted on the COCO and Cityscapes datasets validate the efficacy of our proposed method. Our approach establishes a new benchmark for SSIS, outperforming previous methods.
Specifically, our DG achieves 22.29\%, 31.47\%, and 35.14\% mAP for 1\%, 5\%, and 10\% labeled data on the COCO dataset, respectively.
    
\end{abstract}

\section{Introduction}
    Instance segmentation aims to detect each object and represent it with a segmentation mask~\cite{Li2022A}. Recently, pursuing advanced instance segmentation techniques has led to significant strides in understanding complex visual scenes~\cite{bolya2019yolact,bolya2020yolact++, liu2021swin, Cheng2022maskedattention,wang2023internimage}. However, instance segmentation methods heavily rely on fully annotated datasets, which are often laborious and expensive to create. Among these, semi-supervised learning emerges as a promising paradigm, particularly in scenarios constrained by limited labeled data. 
    Existing semi-supervised instance segmentation pipelines typically generate pseudo labels from unlabeled images and train the model with the labeled image. Therefore, how to use unlabeled data more effectively is a significant challenge of semi-supervised instance segmentation, and the difficulty of accurately encoding the geometric information in the image makes this task more difficult. An insight is that if we can get more comprehensive and rich features from the image, we can get more accurate pseudo tags, and then guide the model to learn more effectively. Generally, semi-supervised instance segmentation frameworks use RGB images to train models~\cite{xu2021endtoend, wang2022noisy, Hu2023pseudolabel, berrada2023guided, Yang2022RevisitingWC, 
 jaemin2023switching}. Although these methods are effective to some extent, they often fail to meet the requirements in scenes where understanding the depth and geometry of the scene is crucial. 

\begin{figure}[t]
  \centering 
  \includegraphics[width=1\linewidth]{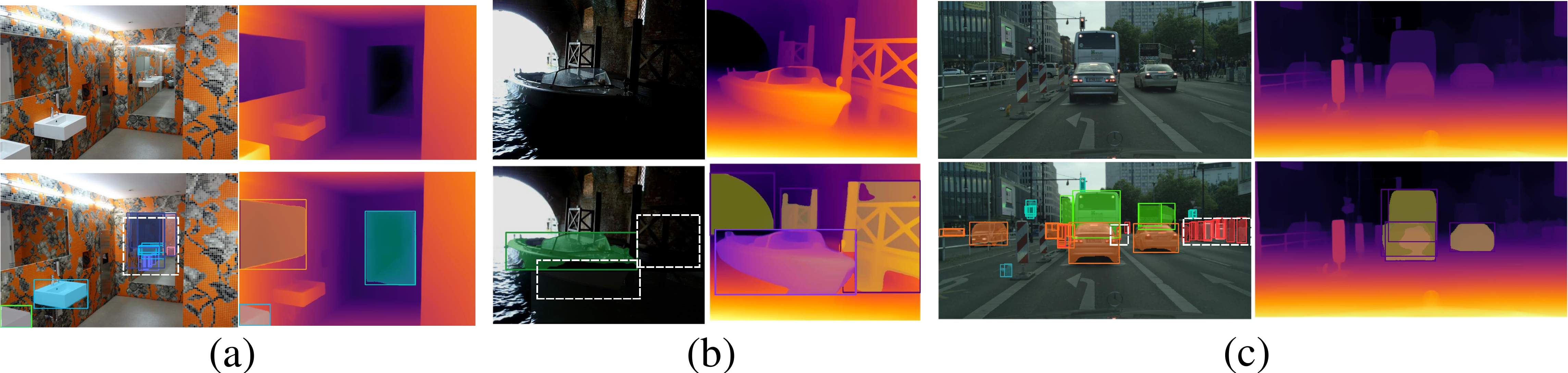}
  \caption{To illustrate the characteristics of depth maps and their complementary role in capturing spatial information from RGB images, we present segmentation results obtained from RGB images and depth maps. (a) and (b) are from COCO~\cite{lin2014microsoft}, (c) is from Cityscapes~\cite{Cordts2016Cityscapes}. (a) The depth map neglects the images reflected in the mirror. (b) The depth map predicts the dark areas of the boat hull and the fence. (c) The depth map is more inclined towards the segmentation of foreground vehicles, and the result treats each car as a whole entity.} 
  \label{fig1} 
\end{figure}

     Studies have shown that combining RGB images with depth data can bring a better segmentation effect~\cite{Zhang2023Overview}. Specifically, depth maps compensate for spatial information missing in RGB images. Compared to RGB images, depth data focuses more on spatial positional information while to some extent disregarding texture information. This characteristic often results in superior segmentation outcomes on certain images when using depth data. As illustrated in Figure ~\ref{fig1}, the top row presents the original RGB image and its corresponding depth map, while the bottom row displays the segmentation results corresponding to the top row. Attention is drawn to the areas outlined by white dashed lines in the figure, where depth data notably exhibits less susceptibility to texture features and environmental conditions. It prioritizes foreground information and avoids over-segmentation of objects.
     
    In this paper, we present a new semi-supervised instance segmentation framework, that introduces monocular depth estimation maps into the field of semi-supervised instance segmentation. It utilizes geometric pseudo labels derived from depth maps and traditional RGB-based pseudo labels to improve the learning efficiency and accuracy of student models. Specifically, the pseudo labels generated from depth maps and RGB images guide the prediction of the student model, while extracting features from depth maps and integrating them into the student model to ensure effective utilization of depth information. In addition, recognizing the dynamic nature of learning in semi-supervised environments, we introduce an adaptive optimization strategy. On the one hand, improving the convergence speed and mitigates the influence of depth maps on model accuracy during the initial stages of training, on the other hand, it fine-tunes the weight ratio between RGB and depth image supervision, thereby enabling a flexible learning process that adapts to the model's evolving requirements. This adaptability plays a crucial role in optimizing performance across various stages of the learning process.
    
We conduct extensive experiments to investigate the characteristics of DG.
Specifically, we demonstrate the superiority of DG under the semi-supervised instance segmentation protocol using public benchmarks, including COCO~\cite{lin2014microsoft} and Cityscapes~\cite{Cordts2016Cityscapes}.
On the COCO dataset, DG achieved 22.29\%, 31.47\%, and 35.14\% AP using labeled data settings of 1\%, 5\%, and 10\%, with 5\% setting exceeding the current SOTA by 1.57\%.
%
To our knowledge, DG is the first framework to introduce depth maps into semi-supervised instance segmentation. Given the effectiveness and conceptual simplicity, we hope DG can serve as a strong baseline for future research in semi-supervised instance segmentation.

\section{Related Work}

\subsection{Semi-supervised Learning.}
The core issue in SSL lies in how to design reasonable and effective supervision signals for unlabeled data. The main methods can be classified into two categories: pseudo-label-based~\cite{scudder1965probability, Lee2013pseudolabel, pham2021meta,rosenberg2005semi,xie2020self,zoph2020rethinking} and consistency-regularization-based~\cite{bachman2014learning,rasmus2015semi,laine2016temporal,sajjadi2016regularization,xie2020unsupervised,berthelot2019remixmatch,dai2017good,berthelot2019mixmatch,gong2021alphamatch} methods. 
Specifically, pseudo-label-based methods leverage pre-trained models to generate annotations for the unlabeled images to train the model and then combine them with manually labeled data for further re-training. 
For another thing, consistency regularization holds the assumption that the prediction of an unlabeled example should be invariant to different forms of perturbations. incorporate various data augmentation techniques to generate different inputs for one image and enforce consistency between these inputs during training.
Among them, FixMatch~\cite{sohn2020fixmatch} combines the consistency regularization-based techniques with a pseudo-label-based framework by applying a strong-weak data augmentation pipeline to input images and enforcing consistency between the augmented images. 
In this work, we follow the pseudo-label-based methods and also use strong-weak data augmentation during training in DG.

\subsection{Semi-supervised instance segmentation}
Contrary to object detection and semantic segmentation, instance segmentation in the semi-supervised setting has received limited attention. Among the few studies:
Noisy Boundaries~\cite{wang2022noisy} was the pioneer in formally proposing the semi-supervised instance segmentation task. It utilizes the Teacher-Student paradigm where the student network learns a noise-tolerant mask head for boundary prediction by leveraging low-resolution features.
Polite Teacher~\cite{filipiak2022Polite} focuses on modern anchor-free detectors and incorporates mask scoring~\cite{Huang2019Mask} for pseudo-mask thresholding.
PAIS~\cite{Hu2023pseudolabel} explores the use of pseudo-labels with low confidence in both box-free and box-dependent instance segmentation contexts.
GuidedDistill~\cite{berrada2023guided} introduces a specific form of knowledge distillation (KD) and develops an improved burn-in stage before commencing the main semi-supervised training loop.

\subsection{Monocular depth estimation (MDE)}
Early monocular depth estimation (MDE)~\cite{hoiem2007recovering,liu2008sift,Saxena2008make3d} initially depended on handcrafted features and traditional computer vision techniques. Recently, deep learning-based methods have significantly advanced, enabling direct regression of scene depth from input images~\cite{Eigen2014Depth}.
ZoeDeph~\cite{Bhat2023zoedepth} introduces a model that combines relative and metric depth estimation, achieving excellent generalization performance while preserving metric scale accuracy.
A landmark development, MiDaS~\cite{Birkl2023midasv31}, employs an affine-invariant loss to accommodate varying depth scales and shifts across different datasets, facilitating effective multi-dataset joint training.
Depth Anything~\cite{Yang2024Depth} expands the dataset by developing a data engine that collects and automatically annotates large-scale unlabeled data. This approach establishes a more challenging optimization target and auxiliary supervision, building a simple yet powerful foundation model.

\section{Method}

\subsection{Preliminary}
In semi-supervised instance segmentation, the dataset consists of two parts: unlabeled images $\mathcal{D}^u=\left \{ x_i^u \right \} $, labeled images $\mathcal{D}^l=\left \{ x_i^l, y_i \right \} $, where $x_i$ is an image，$y_i=\left \{ (y_i^k, c_i^k) \right \}_{k=1,\dots,n}$ represents the associated ground-truth instances defined by $\mathit{n}$ binary masks $y^k_i$ and class indices $c^k_i$.
The purpose of this setting is to jointly use $\mathcal{D}^u$ and $\mathcal{D}^l$ to maximize the performance of the instance segmentation model.
The teacher-student framework is widely adopted in semi-supervised learning, comprising a teacher model and a student model. 
The pre-trained teacher model generates pseudo-labels for unlabeled images, which serve as ground truth to train the student model, thus functioning as a form of self-supervised learning. 
In semi-supervised instance segmentation, the loss function can be decomposed into two terms for labeled and unlabeled images: 
\begin{equation}
L_{\text{semi}} = \lambda_lL_l + \lambda_uL_u 
\label{eq_total}
\end{equation}
where $\lambda_l$ and $\lambda_u$are hyper-parameters for balancing the loss terms. 

As Mask2Former~\cite{Cheng2022maskedattention} is a powerful instance segmentation model, we take it as our meta-model architecture.
For label image $x_i^l$, we match predictions from the student model to ground-truth instances $y_i$ by constructing a cost matrix for bipartite matching, similar to DETR~\cite{Carion2020EndtoEndOD}.
%
%
%
After matching, we use a weighted sum of cross-entropy and Dice loss for the masks, and cross-entropy loss for the class predictions:
%
\begin{equation}
\mathcal{L}_{l}^{i}=\frac{1}{n} \sum_{k=1}^{n} l_{C E}\left(p_{i}, y_{i}^{k}\right)+\lambda_{D} l_{D}\left(p_{i}, y_{i}^{k}\right)+\lambda_{C} l_{C}\left(\hat{c}_{i}, c_{i}^{k}\right) 
\end{equation}
\begin{equation}
\mathcal{L}_{l}=\frac{1}{|B|} \sum_{i \in B}\mathcal{L}_{s}^{i},
\end{equation}
where $p_i$ is the student model’s prediction, $l_{CE}$ is the cross-entropy loss for the class predictions, $l_C$ binary cross-entropy and $l_D$the dice loss function, $\lambda_C$ and $\lambda_D$ are scaling parameters, and $\mathit{B}$ is the training batch. For the unsupervised loss $L_u$, we follow the same procedure as the supervised setting, just replacing $y_i$ with pseudo ground-truth instances $y^{t}_i$ from the teacher.
And explain $L_u$ in section~\ref{sec32}
The student model is updated directly using the gradient descent algorithm， while the teacher model’s parameters $\theta_t$ are updated with the Exponential Moving Average~(EMA) of the student model’s parameters $\theta_s$ as:
\begin{equation}
\theta_{t}^{n+1}\leftarrow\alpha \cdot \theta_{t}^{n}+(1-\alpha) \cdot \theta_{s}^{n+1}
\end{equation}

\textbf{Training methodology.}
Referring to the training strategy of GuidedDistill\cite{berrada2023guided}, we divide the training into three stages, as shown in the right of Figure~\ref{fig1},
\textit{Step 1}. Teacher training stage: the teacher model, parameterized by $\theta_t$, is trained on labeled data $\mathcal{D}^l$. 
\textit{Step 2}. Student training stage: the student model $\theta_s$ is initialized randomly and trained on labeled $\mathcal{D}^l$ and unlabeled $\mathcal{D}^u$ data using pseudo-labels provided by a pre-trained teacher model. During this phase, the teacher model remains unaltered. 
\textit{Step 3}. Teacher Exponential Moving Average (EMA) updating stage: initially, the weights of the student are copied to the teacher, and then the student is trained on labeled $\mathcal{D}^l$ and unlabeled $\mathcal{D}^u$ data as before. During this phase, the teacher model is updated using the EMA of the student weights.

\begin{figure}[t]
    \centering
    \includegraphics[width=1\linewidth]{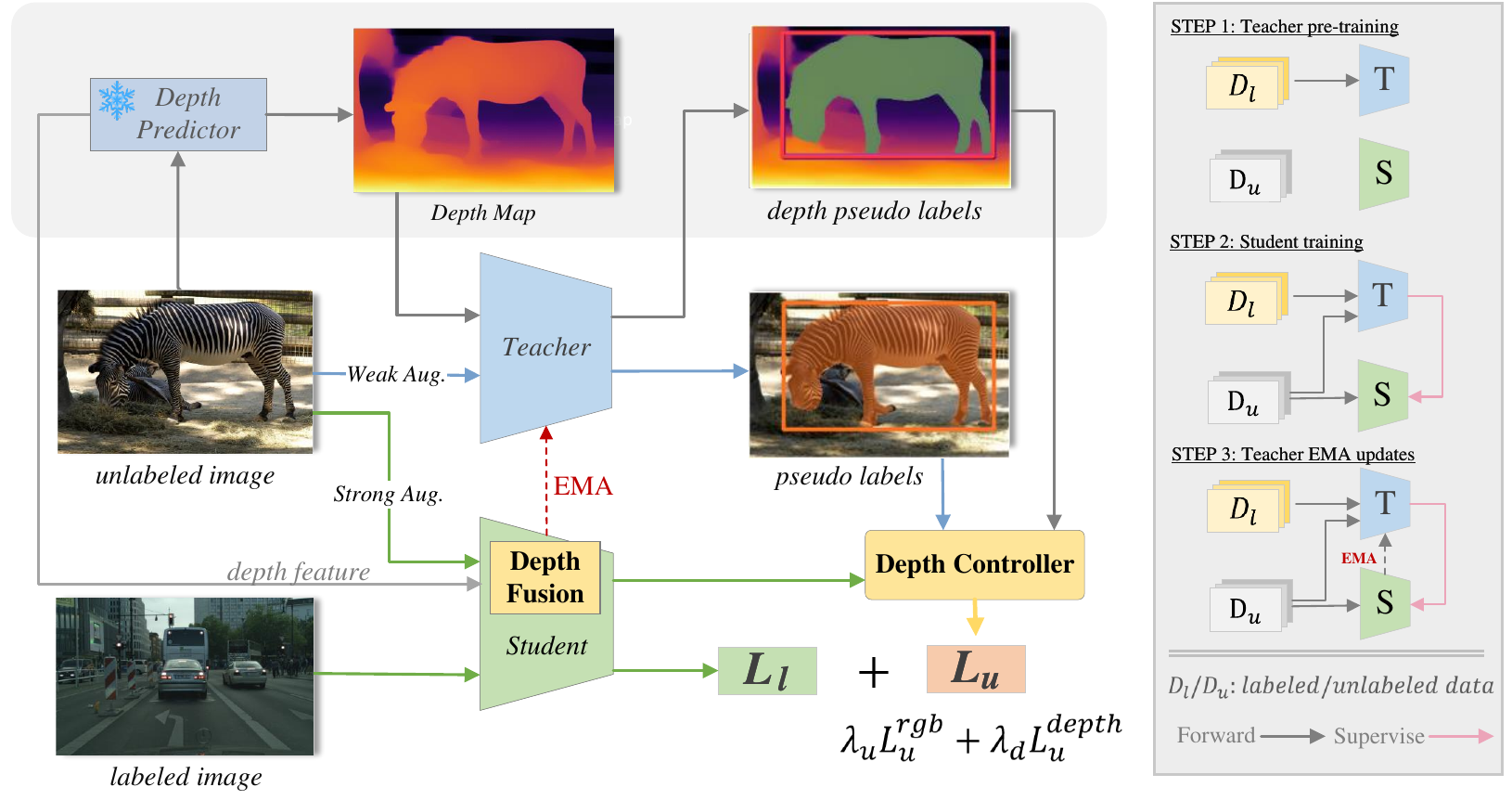}\\
    \caption{\textbf{Framework and training methodology of Depth-Guided (DG).} The focus of DG is on the use of depth information.
    To achieve this, we incorporate a frozen pre-trained depth detector into our framework and propose \textbf{Depth Fusion (DF)} and \textbf{Depth Controller (DC)} to ensure effective utilization of depth information. Specifically,
    Depth map and RGB as inputs for the teacher model to obtain the pseudo-labels that serve as supervision for the student model.
    Meanwhile, Features from the depth detector decoder are fused with the backbone of the student model.
    Last, The Depth Controller module adaptively adjusts the weight of the depth map in the unsupervised loss.
    }
    \label{fig2}
    \vspace{-1.0em}
\end{figure}

\subsection{Depth Map Supervision}
\label{sec32}
To incorporate depth maps into the semi-supervised teacher-student framework, we introduce a novel and effective approach: Depth-Aware Fusion (DG). The key innovation compared to traditional semi-supervised teacher-student frameworks is the inclusion of a depth map, predicted by a pre-trained depth detector~\cite{Yang2024Depth}, for unlabeled images, as illustrated in the upper left corner of Figure~\ref{fig1}. 
%
%
%
Specifically, an unlabeled RGB image is first fed into the depth detector, which outputs a monocular depth estimation (MDE) image of the same size as the original image.
Next, the grayscale MDE image is transformed into a three-channel depth map RGB image by applying a color map.
%
%
Finally, both the unlabeled RGB images and the depth map RGB images are input into the teacher model to generate pseudo-labels. 
These two sets of pseudo-labels collectively serve as the ground truth for the student model's predictions, resulting in losses $L^{rgb}_u$ and $L^{depth}_u$. We replace loss $L_u$ in Eq.~\ref{eq_total} with a linear combination of the two as follows:
\begin{equation}
\mathcal{L}{u}= \lambda_lL_u^{rgb}+\lambda_dL_u^{depth}
\end{equation}
where $\lambda_l$ is the hyper-parameter for the unlabeled RGB image loss, and $\lambda_d$ is the weight for the depth map RGB image loss. The calculation of $\lambda_d$ will be introduced in Sec.~\ref{sec34}.

\begin{figure}[t!]
    \centering
    \includegraphics[width=0.9\linewidth]{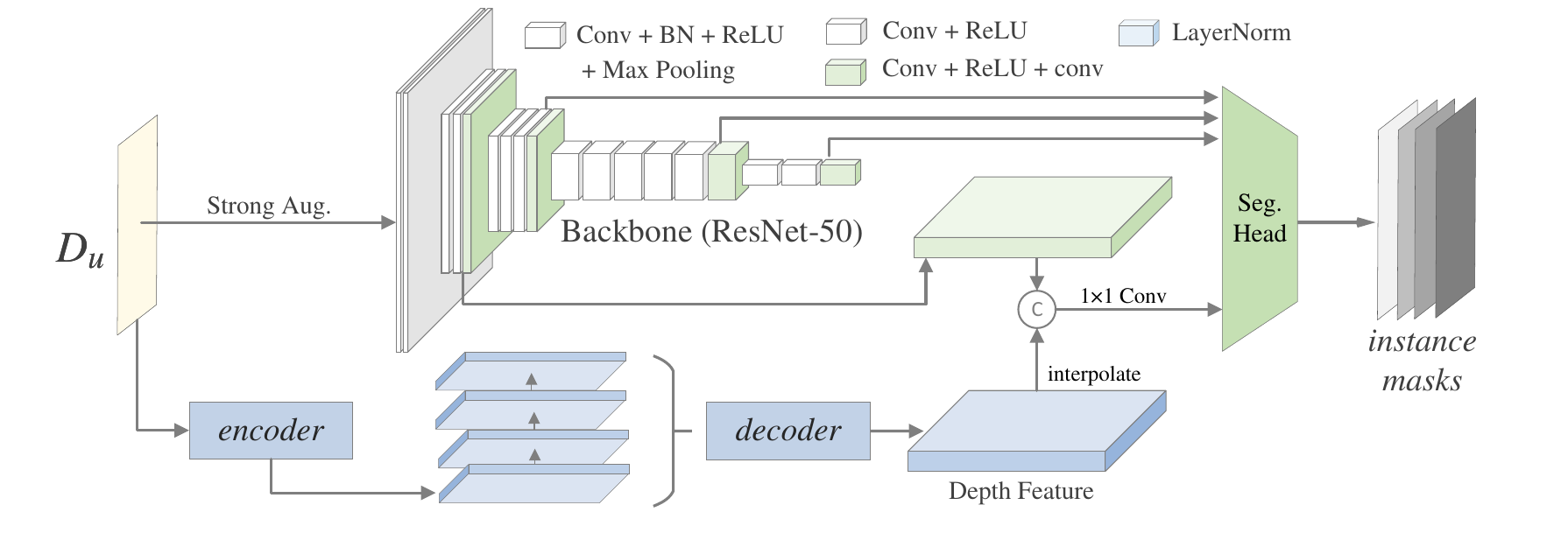}\\
    \caption{Detailed of the Depth Fusion(DF). Unlabeled images($D_u$) input into the student model and the depth estimation model. The features from the student's backbone are fused with the features extracted by the depth estimation model. Finally, these fused features are put into the segmentation head to output the predicted instance masks.
    }
    \label{fig3_DF_detail}
    \vspace{-1em}
\end{figure}

\subsection{Depth Fusion}
To better leverage depth information and enhance the learning of depth features, we propose a Feature Fusion Module that integrates depth features extracted from unlabeled RGB images into the feature space of the student model's backbone network. 
As illustrated in Figure~\ref{fig3_DF_detail}, the depth detector comprises an encoder and a decoder. 
The encoder functions as a versatile multitask encoder capable of generating universal visual features suitable for various image distributions and tasks.
The decoder receives multi-scale feature maps from the encoder, undergoes multiple rearrangements and MLP mappings, merges all scale features to obtain a final feature map, and ultimately produces the output depth feature through convolution. 
%
We use the depth feature generated after multi-layer feature fusion in the decoder of depth detector and input it into the student model. 
During each training iteration, the depth feature is fused with the feature extracted by the student model's backbone network. The fusion is expressed as follows:
\begin{equation}
\mathcal{F}_{fused} = \mathit{Conv}_{(1×1)}\big(concat(\mathit{interp}(\mathcal{F}_\mathit{d}), \mathcal{F}_\mathit{s})\big)
\end{equation}
Where $\mathcal{F}_\mathit{d}$ represents the depth feature and  $\mathcal{F}_\mathit{s}$ represents the feature extracted by the student model's backbone network. $\mathit{interp}$ is an interpolate function used for feature alignment.

After obtaining the fused features $\mathcal{F}_{fused}$, they are input into the segmentation head of the student model to generate the final prediction. It is important to note that the feature fusion operation begins simultaneously with the EMA updating operation of the teacher model. Only after the initiation of EMA updating can the weights from the fusion of depth features be propagated to the teacher model, thereby guiding it to better learn depth features.

\subsection{Depth Controller}
\label{sec34}
Although incorporating depth maps into the semi-supervised training process, due to the significant disparity between the depth map and the original image, the model may exhibit disadvantages such as delayed start and slow convergence in the early stages of training. Moreover, controlling the weight relationship between $L_u^{rgb}$ and $L_u^{depth}$ is challenging. To expedite the model's convergence speed and better control the weight of the unsupervised loss, we introduce an adaptive optimization strategy called the depth controller, which dynamically balances the loss weights between RGB images and depth maps.

Inspired by SoftMatch\cite{Chen2023SoftMatchAT}, we adopt the concept of SoftMatch to decide whether to incorporate less reliable pseudo-labels into the training process based on their quality. Similarly, during each training iteration, we determine the weight of the depth map loss based on the quality of pseudo-labels generated by the teacher model.
Specifically, the teacher model outputs instance-level pseudo-labels and corresponding class confidences in each iteration. We select the instance with the highest confidences $\max(\mathbf{p})$ and fit a \textit{dynamic and truncated Gaussian distribution}. We compute the mean $µ_t$ and variance $\sigma_t$ as follows:
\begin{equation}
\begin{aligned}
\hat{\mu}_{b} & =\hat{\mathbb{E}}_{B_{U}}[\max(\mathbf{p})]=\frac{1}{B_{U}} \sum_{i=1}^{B_{U}} \max(\mathbf{p}_i),
\\
\hat{\sigma}_{b}^{2} & =\hat{\operatorname{Var}}_{B_{U}}[\max(\mathbf{p})]=\frac{1}{B_{U}} \sum_{i=1}^{B_{U}}(\max(\mathbf{p}_i)-\hat{\mu}_{b})^{2}.
\end{aligned}
\end{equation}
where $\hat{\mu}_{b}, \hat{\sigma}_{b}\in \mathbb{R}$. We then aggregate the batch statistics for a more stable estimation, using Exponential Moving Average (EMA) with a momentum m over previous batches, and employ unbiased variance for EMA, initializing $\hat\mu_0$ as $\frac{1}{c}$ and $\hat\sigma_0^2$ as 1.0. The estimated mean $\hat\mu_t$ and variance $\sigma^2_t$ are substituted into the following formula to calculate sample weights:
\begin{equation}
\lambda(\mathbf{p})=\left\{\begin{array}{ll}
\lambda_{\max } \exp \left(-\frac{\left(\max (\mathbf{p})-\mu_{t}\right)^{2}}{2 \sigma_{t}^{2}}\right), & \text { if } \max (\mathbf{p})<\mu_{t}, \\
\lambda_{\max }, & \text { otherwise. }
\end{array}\right.
\end{equation}
which is also a \textit{truncated Gaussian function} within the range [0, $\lambda_{\max }$], on the confidence $\max(\mathbf{p})$.
Finally, $\lambda_d=\mu_{t}-{\lambda}(\mathbf{p})$, if formula is greater than 0, $\lambda_d$ is equal to this, otherwise it is 0.


\section{Experiment}
In this section, we demonstrate the efficacy of our proposed method by conducting comprehensive comparisons with state-of-the-art approaches on various publicly available benchmarks. Furthermore, we provide additional ablation studies to justify the significance and impact of our method.

\begin{table*}[t!]
\renewcommand{\arraystretch}{1.0}
    \begin{center}
    \caption{
    \label{tab1_compare_coco_AP}Comparison of results on COCO. Results for Mask-RCNN, CenterMask2, Polite Teacher, Noisy Boundaries, PAIS, and GuidedDistillation are taken from ~\cite{wang2022noisy}, ~\cite{Hu2023pseudolabel} and ~\cite{berrada2023guided}. $^\ast$ denotes data from NoisyBoundary. 
    } 
    \vspace{1mm}
    \begin{tabular}{l|ccccc}
    \toprule[1.0pt]
        Method & 1\%  & 2\% & 5\% & 10\% & 100\%  \\ \midrule[0.5pt]
        Mask-RCNN$^\ast$~\cite{maskrcnn}, supervised  & 3.50 & -  & 17.30  & 22.00 & 34.50\\ 
        Mask2Former~\cite{Cheng2022maskedattention}, supervised & 13.50 & 20.00 & 26.00 & 30.50 & \underline{43.50}\\ \midrule[0.5pt] \midrule[0.5pt]
        DD~\cite{Ilija2018Data} & 3.80 & 11.80 & 20.40 & 24.20 & 35.70\\ 
        Noisy Boundaries~\cite{wang2022noisy} & 7.70 & 16.30 & 24.90 & 29.20 & 38.60\\ 
        Polite Teacher~\cite{filipiak2022Polite} & 18.30 & 22.30 & 26.50 & 30.80 & -\\ 
        Mask-RCNN, PAIS~\cite{Hu2023pseudolabel} & 21.12 & - & 29.28 & 31.03 &39.50\\ 
        GuidedDistillation~\cite{berrada2023guided} & \underline{21.50} & \underline{25.30} & \underline{29.9} & \underline{35.00} & -\\ \midrule[0.5pt]
        DG, Ours & \textbf{22.29}\tiny{\color{green}({+8.79})} & \textbf{26.28}\tiny{\color{green}({+6.28})} & \textbf{31.47}\tiny{\color{green}({+5.47})} & \textbf{35.14}\tiny{\color{green}({+4.64})} & \textbf{48.84}\tiny{\color{green}({+5.34})}\\ 
        \bottomrule[1.0pt]
    \end{tabular}
    \end{center}
\end{table*}

\subsection{Experimental setup}
\textbf{Datasets and evaluation metric.} 
\textbf{COCO.} The MS COCO dataset~\cite{lin2014microsoft} is a widely used benchmark, which gathers images of complex everyday scenes containing common objects in their natural context. It has instance segmentation for 118,287 training images. Following ~\cite{wang2022noisy}, we use 1\%, 2\%, 5\%, and 10\% of the images from the training set as labeled data for semi-supervised training. We use the same supervised/unsupervised splits as earlier work. 
\textbf{Cityscapes.} The Cityscapes dataset~\cite{Cordts2016Cityscapes} gathers images in urban driving scenes, which contains 2,975 training images of size 1024 × 2048 labeled with 8 semantic instance categories. We adopt the same valuation setting as~\cite{berrada2023guided}, we train models using 5\%, 10\%, 20\%, and 30\% of the available instance annotations and evaluate using the 500 validation images with public annotations. The different data splits are generated by randomly selecting a random subset from the training images. 
\textbf{evaluation metric.} On both datasets, the instance segmentation results are evaluated by the mask-AP metric, including AP and AP50. 

\textbf{Implementation details.}
We adopt mask2former~\cite{Cheng2022maskedattention} based on ResNet-50~\cite{He2016Deep} as our instance segmentation model and implement our approach in \cite{berrada2023guided}. 
The models are trained using AdamW with a learning rate of 10−4, weight decay of 0.05, and multiplier of 0.1 for the backbone updates, a batch size of 16.
The hyper-parameters are set as follows. 
the unsupervised loss weight $\lambda_u$ = 2,  the class threshold to $\alpha_C$ = 0.7, and the size threshold to $\alpha_S$ = 5. For the loss weights,  $\lambda_D$ = 5 and $\lambda_C$ = 1. We set the EMA decay rate to $\alpha$ = 0.9996. By default, experiments were conducted on a machine with four A100 GPUs each with 80 GB of RAM. 
For COCO and Cityscapes, we train our models for 368k and 90k iterations. The number of iterations used in the Student training stage is as follows. For COCO dataset, the splits of 1\%, 2\%, 5\%, and 10\% is 15k, 20k, 30k and 60k iterations. For Cityscapes dataset, the splits of 5\%, 10\%, 20\%, and 30\% is 15k, 25k, 30k and 35k iterations. 

\textbf{Depth estimation.}
For obtaining depth maps from RGB images, we utilized the depth anything~\cite{Yang2024Depth} model. This model adopts the DINOv2 encoder~\cite{Oquab2023dinov2} for feature extraction and uses the DPT~\cite{dpt} decoder for depth regression. We utilized its pre-trained model, Depth-Anything-Large, which consists of 335.3 M parameters. It is worth noting that this pre-trained model has not been fine-tuned on the dataset used in our work.

\begin{table}[t!]
\renewcommand{\arraystretch}{1.0}
    \begin{center}
    \caption{
    \label{tab3_compare_cityscapes}Comparison of AP on results. Results for Mask-RCNN, CenterMask2, Polite Teacher, Noisy Boundaries, PAIS, and GuidedDistillation are taken from ~\cite{wang2022noisy}, ~\cite{Hu2023pseudolabel} and ~\cite{berrada2023guided}. $^\ast$ denotes data from NoisyBoundary. 
    } 
    \vspace{1mm}
    \setlength{\tabcolsep}{3.5mm}
    \begin{tabular}{l|ccccc}
    \toprule[1.0pt]
        Method & 5\% & 10\% & 20\% & 30\% & 100\%\\ \midrule[0.5pt]
        Mask-RCNN$^\ast$~\cite{maskrcnn},supervised  & 11.30 & 16.40 & 22.60 & 26.6 & - \\
        Mask2Former~\cite{Cheng2022maskedattention},supervised & 12.10 & 18.80 & 27.40 & 29.6 & 33.70\\ \midrule[0.5pt] \midrule[0.5pt]
        DD~\cite{Ilija2018Data} & 13.70 & 19.20 & 24.60 & 27.4 & -\\
        STAC~\cite{sohn2020simple} & 11.90 & 18.20 & 22.90 & 29.00 & -\\
        CSD~\cite{JisooJeong2019ConsistencybasedSL} & 14.10 & 17.90 & 24.60 & 27.5 & - \\
        CCT~\cite{YassineOuali2020SemiSupervisedSS} & 15.20 & 18.60 & 24.70 & 26.50 & - \\
        Dual-branch~\cite{WenfengLuo2020SemisupervisedSS} & 13.90 & 18.90 & 24.00 & 28.90 & - \\
        Ubteacher~\cite{liu2021unbiased} & 16.00 & 20.00 & 27.10 & 28.00 & - \\
        Noisy Boundaries~\cite{wang2022noisy} & 17.10 & 22.10 & 29.00 & 32.40 & 34.70 \\
        Mask-RCNN, PAIS~\cite{Hu2023pseudolabel} & 18.00 & 22.90 & 29.20 & 32.80 & - \\
        GuidedDistillation~\cite{berrada2023guided} & 23.00 & 30.80 & 33.10 & 35.60 & 39.60\\ \midrule[0.5pt] 
        DG, ours & \textbf{23.23} &\textbf{30.85} &\textbf{34.05}  &\textbf{36.66}  &\textbf{39.80} \\
    \bottomrule[1.0pt]
    \end{tabular}
    \end{center}
    \vspace{-1.0em}
\end{table}

\begin{figure}[t]
    \begin{minipage}{0.5\textwidth}
        \vspace{5mm}
        \tabcaption{\label{tab2_compare_coco_AP50}Comparison of results on COCO. The evaluation metric is AP50.$^{\dagger}$ denotes data is supervised and taken from ~\cite{Hu2023pseudolabel}.$^\ast$ denotes the reproduced results in ~\cite{berrada2023guided}}
    \resizebox{6.5cm}{!}{
        \centering
         \begin{tabular}{l|ccc}         
            \toprule[1.0pt]
            Method & 1\% & 5\% & 10\%   \\ \midrule[0.5pt]
            Mask-RCNN$^{\dagger}$~\cite{maskrcnn}  & 19.86 & 37.98  & 45.10 \\ 
            Mask-RCNN, PAIS~\cite{Hu2023pseudolabel} & \underline{36.03} & \underline{47.25} & 49.83 \\ 
            GuidedDistillation$^\ast$~\cite{berrada2023guided} & 35.66 & 46.66 & \underline{52.66} \\ \midrule[0.5pt]
            DG, ours & \textbf{36.60} & \textbf{49.22} & \textbf{54.03} \\ 
            \bottomrule[1.0pt]
        \end{tabular}
    }

    \end{minipage}
    \hspace{0.01in}
        \begin{minipage}[t]{0.45\linewidth}
            \vspace{-18mm}
            \includegraphics[scale=0.36]{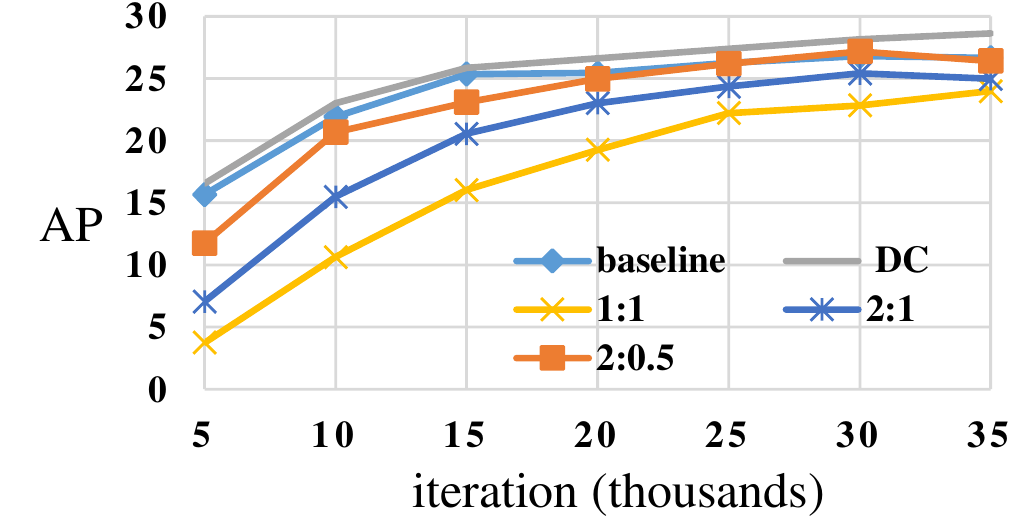}
            \figcaption{Effect of DC on convergence speed. 
         }
        \label{fig3_AP_record}
        \end{minipage}
    \vspace{-1.2em}
\end{figure}


\subsection{Main Results}

\textbf{COCO.}
We compare the results of DG with the current state-of-the-art on the validation set under the ImageNet1k pre-trained ResNet-50 backbone to validate the performance of our method.
The AP and AP50 results are shown in Table ~\ref{tab1_compare_coco_AP} and Table ~\ref{tab2_compare_coco_AP50}. The results demonstrate that DG surpasses the previous methods.
Specifically, we have the following results. When utilizing 1\%, 5\%, and 10\% of labeled images, we achieved average precisions (AP) of 22.29\%, 31.47\%, and 35.14\%, respectively, all of which surpass previous methods. Remarkably, even under the condition of utilizing less labeled data, our approach outperforms methods employing more labeled data, such as achieving an AP of 31.47\% with only 5\% labeled data, surpassing PAIS 10\% labeled data performance. To comprehensively present the results, we provide four supervised baselines: Mask-RCNN, PAIS, CenterMask2, and GuidedDistill.

\textbf{Cityscapes.}
Table ~\ref{tab3_compare_cityscapes} reports comparison results of our method against several existing methods on the Cityscapes validation set. Our method yields improvements of +11.13\%, +12.05\%, +6.65\%, and +7.06\% over the Mask2former(supervised) in the 5\%, 10\%, 20\%, and 30\% label partitions, respectively, demonstrating its effectiveness, particularly in the scarce-label setting. 
Notably, our method outperforms PAIS~\cite{Hu2023pseudolabel} on 5\% partition, despite using only half as many trainable parameters. 
In addition, DG brings less improvement on the Cityscapes dataset than on the COCO dataset, and even overfitting occurs at the late stage of training, which we attribute to the more homogeneous scenarios and fewer samples in the Cityscapes dataset compared to the COCO dataset.


\subsection{Empirical Studies}
In the following, unless specified otherwise, ablations are performed on COCO with 5\% labeled data and a ResNet-50 backbone. Results are reported on the validation set.

\begin{wraptable}{r}{8cm}
\centering
     \caption{
        \label{tab4_ablation_components}Ablation studies on our DG. "DS": Depth Map Supervision. "DF": Depth Feature Fusion. "DC": Depth Controller. Improvements over the baseline are highlighted in \textcolor{blue}{blue}.} 
\begin{tabular}{ccc|cc}
    \toprule[1.0pt]
        DS & DF & DC & AP    & AP50  \\ \midrule[0.5pt]
           &    &    & 29.90 & 46.66\\
        \checkmark  &    &    & 30.76$( \textcolor{blue}{0.86\uparrow})$ & 48.52$(\textcolor{blue}{1.86\uparrow})$ \\
        \checkmark  & \checkmark  &    & 31.11$( \textcolor{blue}{1.21\uparrow})$ & 49.00$( \textcolor{blue}{2.34\uparrow})$ \\
        \checkmark  &    & \checkmark  & 30.95$( \textcolor{blue}{1.05\uparrow})$ & 48.62$( \textcolor{blue}{1.96\uparrow})$\\
        \checkmark  & \checkmark  & \checkmark  & 31.47$( \textcolor{blue}{1.57\uparrow})$ & 49.22$( \textcolor{blue}{2.56\uparrow})$ \\
    \bottomrule[1.0pt] 
    \end{tabular}
\end{wraptable}

\textbf{Effectiveness of different components of DG.}
We first investigate the effectiveness of each component of DG in Table ~\ref{tab4_ablation_components}.
We set the baseline as the performance of GuidedDistill using 5\% labeled images on the COCO dataset, demonstrating the efficacy of our proposed three components in enhancing the performance of SSIS, as measured by AP and AP50.
Utilizing Depth Map Supervision (DS) leads to improvements of 0.86\% and 1.86\%.
Subsequently, incorporating Depth Feature Fusion (DF) and Depth Controller (DC) results in improvements of AP by 1.21\% and 1.05\% respectively, and AP50 by 2.34\% and 1.96\% respectively. These improvements surpass those achieved by DS alone, indicating that the introduction of DF and DC can further improve the model and achieve better performance.
Lastly, integrating both components can further improve each individual component and achieve the best performance, with improvements of 1.57\% and 2.56\%.


\textbf{Depth controller improves convergence speed.}
To illustrate how Depth Controller overcomes the drawbacks of low initial value and slow convergence, because of the use of depth maps, we present AP records under different settings in Figure ~\ref{fig3_AP_record}. Here, "baseline" denotes the AP performance of GuidedDistill using 5\% labeled images on the COCO dataset, "DC" represents 
\begin{wrapfigure}{r}{7cm}
\vspace{-0.5em}
    \centering
    \includegraphics[width=1\linewidth]{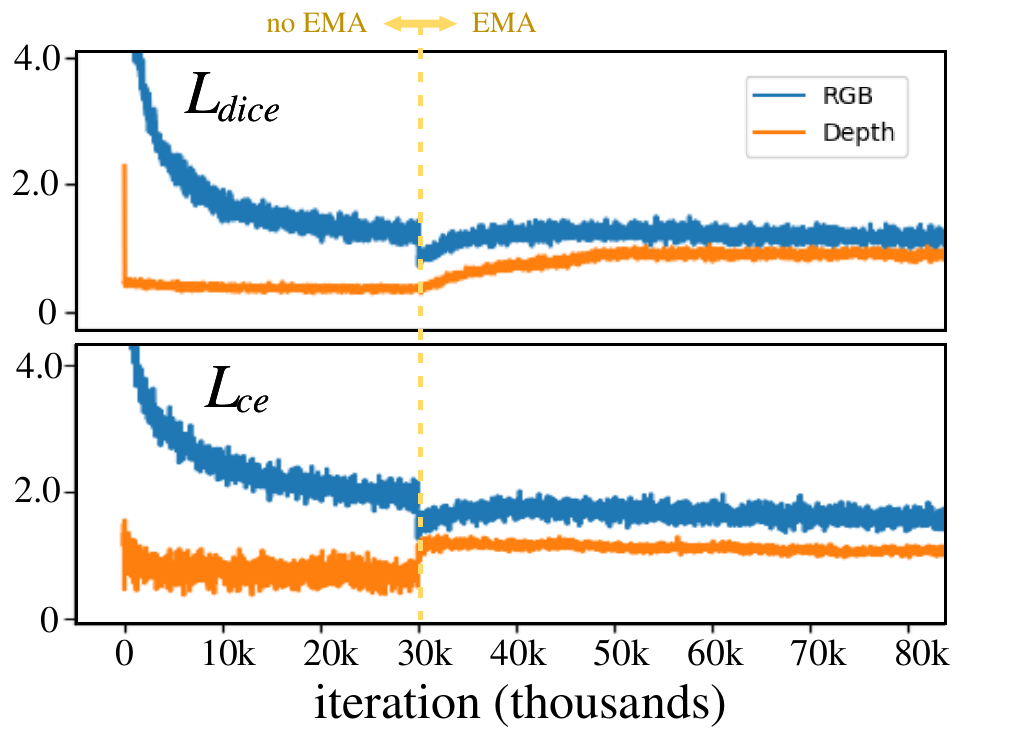}\\
    \caption{Unsupervised Loss record. After iteration over 30k, the teacher model starts the EMA update.}
    \label{fig5_loss_record}
\end{wrapfigure}
the performance with the inclusion of DC, and "1:1", "2:0.5", "2:1" denote different ratios of $\lambda_u:\lambda_d$.
Experimental results indicate that the use of the DC achieves higher initial values and surpasses all other settings simultaneously.
%
It is important to note that the complete AP records are not provided in the figure, and the results under the "2:0.5" setting will surpass the baseline in later stages of training, as shown in Table ~\ref{tab5_ablation_lambda_d}. 
As long as depth information is integrated properly, sooner or later, it can enhance the model's performance.


\begin{table}[t!]
    \renewcommand{\arraystretch}{1.0}
    \begin{center}
        \caption{
            \label{tab5_ablation_lambda_d}Ablations on the loss weight $\lambda_d$. "DC": use depth controller.} 
        \setlength{\tabcolsep}{3.0mm}
        \vspace{1mm}
        \begin{tabular}{c|ccccc|c}
        \toprule[1.0pt]
        $\lambda_d$ & 0 & 0.25 & 0.5 & 0.75 & 1   & DC \\ \midrule[0.5pt]
        AP   & 29.90 & 28.97 & 30.76 & 30.20 & 27.71 & \textbf{31.47}            \\
        AP50 & 46.66 & 46.25 & 48.52 & 48.05 & 44.60 & \textbf{49.22}     \\
        \bottomrule[1.0pt]
    \end{tabular}
    \end{center}
    \vspace{-1.2em}
\end{table}

\textbf{Unsupervised loss weight.}
We also examined the DG with different values of unsupervised loss weight $\lambda_d$ and reported both AP and AP50. In Table ~\ref{tab5_ablation_lambda_d}, $\lambda_d=0$ as the baseline, which is ~\cite{berrada2023guided} on the COCO dataset using 5\% labeled images. we find it optimal to use $\lambda_u$ = 0.5. However, when using the Depth Controller to adjust $\lambda_d$ adaptively, we achieved higher AP and AP50 scores.

\textbf{Unsupervised loss record.}
%
%
In Figure ~\ref{fig5_loss_record} displays the record of $L_u^{rgb}$ and $L_u^{depth}$. The top shows the Dice loss for the masks, the bottom shows the cross-entropy loss for the class. It can be observed that the $L_u^{depth}$ is more stable than the $L_u^{rgb}$. A significant increase occurs when the teacher model begins EMA updates, which effectively complements the ongoing decline in $L_u^{rgb}$.

\begin{figure}[t]
    \centering
    \includegraphics[width=1.0\linewidth]{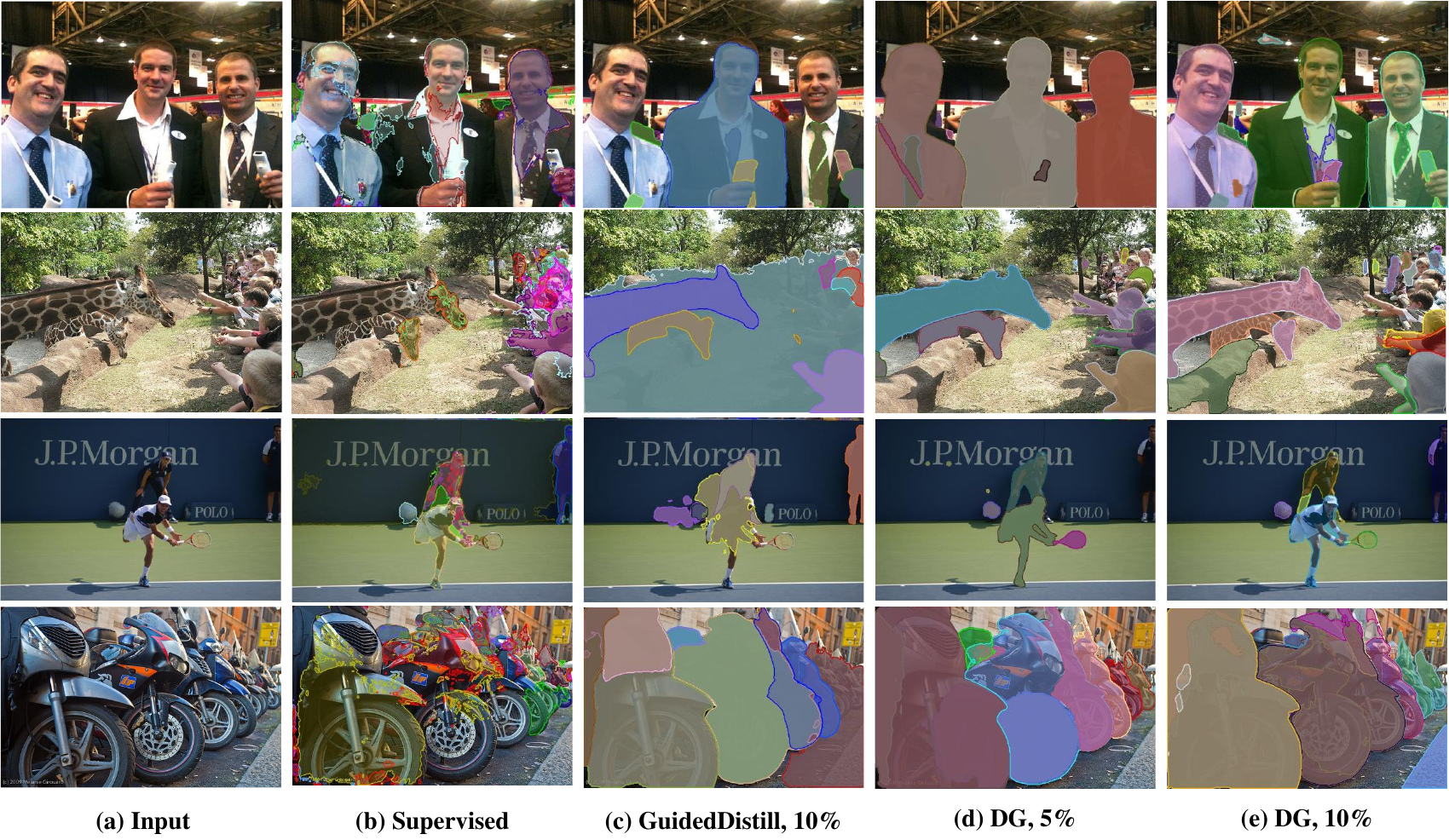}
    \caption{Visualization of instance segmentation results on COCO validation set. 
    All methods are evaluated with Mask2former on the ResNet-50.
    (b)(d) use 5\% labeled data. (c)(d) use 10\% labeled data.}
    \label{fig6_vis}
    \vspace{-1.5em}
\end{figure}

\textbf{Qualitative Results.}
We visualize the instance segmentation results to compare our method with the baseline in Figure~\ref{fig6_vis}. Column (b) displays the output from a supervised-only approach trained with a limited amount of labeled data, producing the least credible segmentation results.
Columns (c) and (e) represent the visualization of the top ten most confident masks for Guided Distillation and DG, respectively, when using 10\% labeled data. Our DG manifests stronger instance segmentation capability, offering clearer boundaries for the segmented instances and better differentiation between the foreground and background.
Column (d) demonstrates the visualized results of DG using 5\% labeled data. Despite employing fewer labeled data compared with column (c), it still achieves comparable accuracy results.
Please refer to the following supplementary material for more qualitative results.

\section{Conclusion}
In this paper, we proposed the Depth-Guided Semi-Supervised Instance Segmentation (DG-SSIS) framework introduces a novel integration of depth maps with RGB data, enhancing the accuracy and robustness of instance segmentation in semi-supervised settings. By deploying Depth Feature Fusion and the Depth Controller, our method capably adjusts and balances the impact of depth information, thus overcoming the challenges associated with its integration. 
This work sets a new benchmark for semi-supervised instance segmentation and lets SSIS have a deeper understanding of image depth and geometry, offering novel insights into integrating geometric information into segment models.

\textbf{Broader Impact and Limitations.}
This research effectively integrates depth information into semi-supervised learning, leveraging complementary data to enhance the performance. 
To our knowledge, DG is the first framework to use monocular depth estimation images in semi-supervised instance segmentation. We hope DG can serve as a strong baseline for future research in semi-supervised instance segmentation.
We use our approach on GD, but our components can also be integrated within other frameworks.
In addition, in this paper, only ResNet-50 was used as the backbone for extracting features, and a larger backbone would have resulted in superior performance.


\bibliography{Styles/reference}

\begin{titlepage}
\appendix

\noindent\textbf{\Large Supplementary Material}
\section{Teacher model after step 1}

Our training is divided into 3 stages: Teacher pre-training, student training, and Teacher EMA update, the latter two stages have been explained in detail above. Here, we add to the first stage, in Table~\ref{tab6_appendix_step1}, which shows the AP and AP50 achieved by pre-training with different percentages of labeled data on COCO and Cityscapes datasets.
\vspace{-1.0em}
\captionsetup[subfloat]{
    position=bottom, 
}

\begin{table}[h]
\caption{Teacher model's AP, AP50 and iteration after pre-train.}
\vspace{2mm}
\centering
    \scalebox{0.8}{  
    \subfloat[COCO]{
        \begin{tabular}{c|ccccc}
        \toprule[1.0pt]
        Labels used& 1\% & 2\% & 5\% & 10\% & 100\% \\ \midrule[0.5pt]
        AP   &13.36 &9.39 &25.8 &30.05 &39.21\\
        AP50 &23.31 &32.12 &41.92 &47.59 &60.4\\ \midrule[0.5pt]
        iteration &30k &35k &50k &100k &195k\\
        \bottomrule[1.0pt]
        \end{tabular}
        \label{coco_step1}
    }}
    \scalebox{0.8}{  
    \subfloat[Cityscapes]{
        \begin{tabular}{c|ccccc}
        \toprule[1.0pt]
        Labels used& 5\% & 10\% & 20\% & 30\% & 100\% \\ \midrule[0.5pt]
        AP   &15.74 &22.74 &29.71 &30.68 &37.69\\
        AP50 &32.05 &41.85 &52.23 &54.5 &63.57\\  \midrule[0.5pt]
        iteration &15k &30k &65k &50k &90k\\
        \bottomrule[1.0pt]
        \end{tabular}
    \label{cityscapes_step1}
    }}
\label{tab6_appendix_step1}
\vspace{-2.0em}
\end{table}

\section{Future Works}
Recently, some work has used a larger backbone in the framework of semi-supervised instance segmentation, such as Swin~\cite{liu2021swin}, DINOv2~\cite{Oquab2023dinov2}, and Deit~\cite{szegedy2015going}, showing higher performance than Resnet~\cite{He2016Deep}.
On the other hand, beyond depth maps, various other types of imagery, including heat maps, edge maps, and corner maps, can be incorporated into semi-supervised learning frameworks. 
Moreover, we posit that models trained on specific image domains exhibit superior performance in relevant contexts. For instance, the integration of depth information can improve the accuracy of datasets with insignificant depth features, such as nighttime and camouflage image datasets. 
In the future, we will try larger backbones, and more image types, and test on more types of datasets.
Finally, our proposed depth feature fusion methodology remains open to optimization. Exploring fusion across multiple layers or scales holds promise for yielding improved outcomes. In the future work, we will also try more.

\section{More Qualitative Results}
We show more visualization results in Figure~\ref{fig7_vis}.
All methods are evaluated with Mask2former on the ResNet-50.
Divided into two columns, each from left to right, in order, is the input image, the result of our DG trained with 5\% labeled data, and the result of GuidedDistill trained with 10\% labeled data.

\begin{figure}[h]
    \centering
    \includegraphics[width=0.9\linewidth]{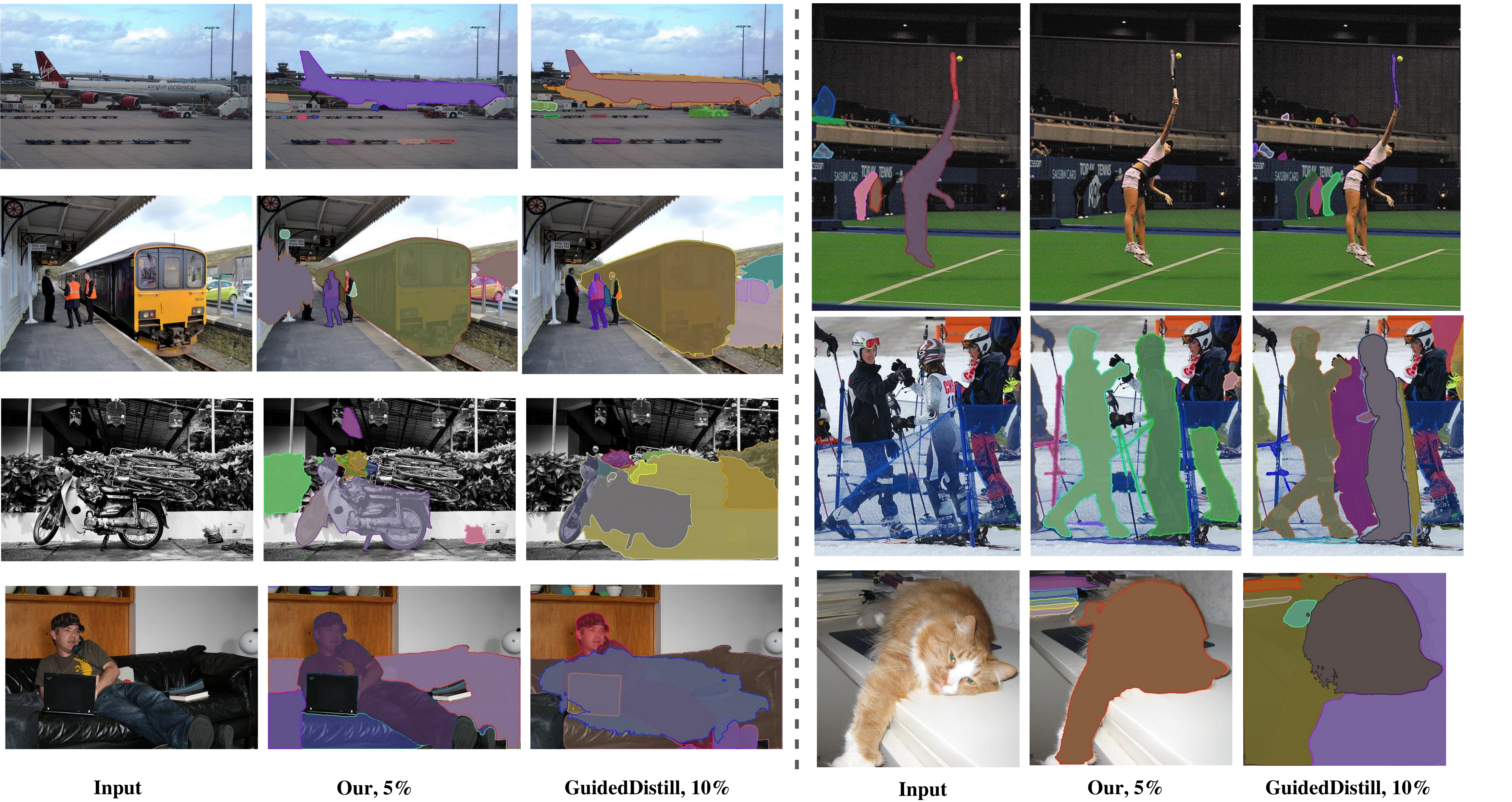}
    \caption{Visualization of instance segmentation results on COCO validation set.}
    \label{fig7_vis}
\end{figure}

\end{titlepage}

\newpage
\section*{NeurIPS Paper Checklist}

\begin{enumerate}

\item {\bf Claims}
    \item[] Question: Do the main claims made in the abstract and introduction accurately reflect the paper's contributions and scope?
    \item[] Answer: \answerYes{} 
    \item[] Justification: \justificationTODO{}

\item {\bf Limitations}
    \item[] Question: Does the paper discuss the limitations of the work performed by the authors?
    \item[] Answer: \answerYes{} 
    \item[] Justification: \justificationTODO{}

\item {\bf Theory Assumptions and Proofs}
    \item[] Question: For each theoretical result, does the paper provide the full set of assumptions and a complete (and correct) proof?
    \item[] Answer: \answerYes{} 
    \item[] Justification: \justificationTODO{}

    \item {\bf Experimental Result Reproducibility}
    \item[] Question: Does the paper fully disclose all the information needed to reproduce the main experimental results of the paper to the extent that it affects the main claims and/or conclusions of the paper (regardless of whether the code and data are provided or not)?
    \item[] Answer: \answerYes{} 
    \item[] Justification: \justificationTODO{}

\item {\bf Open access to data and code}
    \item[] Question: Does the paper provide open access to the data and code, with sufficient instructions to faithfully reproduce the main experimental results, as described in supplemental material?
    \item[] Answer: \answerYes{} 
    \item[] Justification: \justificationTODO{}

\item {\bf Experimental Setting/Details}
    \item[] Question: Does the paper specify all the training and test details (e.g., data splits, hyperparameters, how they were chosen, type of optimizer, etc.) necessary to understand the results?
    \item[] Answer: \answerYes{} 
    \item[] Justification: \justificationTODO{}

\item {\bf Experiment Statistical Significance}
    \item[] Question: Does the paper report error bars suitably and correctly defined or other appropriate information about the statistical significance of the experiments?
    \item[] Answer: \answerYes{} 
    \item[] Justification: \justificationTODO{}

\item {\bf Experiments Compute Resources}
    \item[] Question: For each experiment, does the paper provide sufficient information on the computer resources (type of compute workers, memory, time of execution) needed to reproduce the experiments?
    \item[] Answer: \answerYes{} 
    \item[] Justification: \justificationTODO{}
    
\item {\bf Code Of Ethics}
    \item[] Question: Does the research conducted in the paper conform, in every respect, with the NeurIPS Code of Ethics \url{https://neurips.cc/public/EthicsGuidelines}?
    \item[] Answer: \answerYes{} 
    \item[] Justification: \justificationTODO{}

\item {\bf Broader Impacts}
    \item[] Question: Does the paper discuss both potential positive societal impacts and negative societal impacts of the work performed?
    \item[] Answer: \answerYes{} 
    \item[] Justification: \justificationTODO{}
    
\item {\bf Safeguards}
    \item[] Question: Does the paper describe safeguards that have been put in place for responsible release of data or models that have a high risk for misuse (e.g., pretrained language models, image generators, or scraped datasets)?
    \item[] Answer: \answerYes{} 
    \item[] Justification: \justificationTODO{}

\item {\bf Licenses for existing assets}
    \item[] Question: Are the creators or original owners of assets (e.g., code, data, models), used in the paper, properly credited and are the license and terms of use explicitly mentioned and properly respected?
    \item[] Answer: \answerYes{} 
    \item[] Justification: \justificationTODO{}

\item {\bf New Assets}
    \item[] Question: Are new assets introduced in the paper well documented and is the documentation provided alongside the assets?
    \item[] Answer: \answerYes{} 
    \item[] Justification: \justificationTODO{}

\item {\bf Crowdsourcing and Research with Human Subjects}
    \item[] Question: For crowdsourcing experiments and research with human subjects, does the paper include the full text of instructions given to participants and screenshots, if applicable, as well as details about compensation (if any)? 
    \item[] Answer: \answerNA{} 
    \item[] Justification: This paper does not involve crowdsourcing nor research with human subjects.

\item {\bf Institutional Review Board (IRB) Approvals or Equivalent for Research with Human Subjects}
    \item[] Question: Does the paper describe potential risks incurred by study participants, whether such risks were disclosed to the subjects, and whether Institutional Review Board (IRB) approvals (or an equivalent approval/review based on the requirements of your country or institution) were obtained?
    \item[] Answer: \answerNA{} 
    \item[] Justification: This paper does not involve crowdsourcing nor research with human subjects.

\end{enumerate}

\end{document}